\documentclass[10pt,twocolumn,letterpaper]{article}

\usepackage{cvpr}
\usepackage{times}
\usepackage{epsfig}
\usepackage{graphicx}
\usepackage{amsmath}
\usepackage{amssymb}
\usepackage{algorithm}
\usepackage{algpseudocode}

\cvprfinalcopy % *** Uncomment this line for the final submission

\def\cvprPaperID{****} % *** Enter the CVPR Paper ID here

\ifcvprfinal\fi
\begin{document}

%%%%%%%%% TITLE
\title{Weighted Multisource Tradaboost}

\author{Jo\~{a}o Antunes\\
Carnegie Mellon University\\
Instituto Superior T\'{e}cnico\\
{\tt\small joaoa@andrew.cmu.edu}
\and
Alexandre Bernardino\\
Institudo Superior T\'{e}cnico\\
{\tt\small alex@isr.tecnico.ulisboa.pt}
\and
Asim Smailagic\\
Carnegie Mellon University\\
{\tt\small asim@cs.cmu.edu}
\and
Daniel Siewiorek\\
Carnegie Mellon University\\
{\tt\small dps@cs.cmu.edu}
}

\maketitle
%\thispagestyle{empty}

%%%%%%%%% ABSTRACT
\begin{abstract}
In this paper we propose an improved method for transfer learning that takes into account the balance between target and source data. This method builds on the state-of-the-art Multisource Tradaboost, but  weighs the importance of each datapoint taking into account the amount of target and source data available. A comparative study is then presented exposing the performance of four transfer learning methods as well as the proposed Weighted Multisource Tradaboost. The experimental results show that the proposed method is able to outperform the base method as the number of target samples increase. These results are promising in the sense that source-target ratio weighing may be a path to improve current methods of transfer learning. However, against the asymptotic conjecture of \cite{torrey}, all transfer learning methods tested in this work get outperformed by a no-transfer SVM for large number on target samples.
\end{abstract}

%%%%%%%%% BODY TEXT
\section{Introduction}

Most machine learning techniques are based on the PAC (Probably Approximately Correct)\cite{tradaboost} model, which states that while operating on a learning problem the samples used for training and the samples that we want to classify follow the same probability distribution. However, this assumption does not hold in a variety of cases. Frequently, the data used for training has become obsolete (e.g. due to changes on how data was collected) or simply that the data available is not enough to train a robust classifier. Insufficient data frequently occurs in classifiers that recognize a high number of classes (e.g. in object recognition systems routinely discriminate between $\approx 10^4$ categories)\cite{svm}. In this case, machine learning techniques give very little guarantees about the generalization error obtained.
Transfer Learning is an approach to address the small dataset challenge. The intuition behind transfer learning is to mimic the way humans learn. The data we acquire from all our senses is stored in our memory along with concepts and inferences we make as to how to categorize this data. This makes it so that any new concept to be assimilated is not learned in isolation. Instead, we consider connections between what we already know and try to apply them to the new concept. The goal of transfer learning is to extract relevant information from data that does not need to come from the same probability distribution as the data to be classified by the final model. The ability to leverage more data during the learning process leads to more robust models since more information is used for training.
In this paper, an improvement on a state-of-the-art transfer learning method is presented: Weighted Multisource Tradaboost.Our proposed approach incorporates the belief that if more
target data is available, the contribution of the source data
used in the model should gradually shift from model defining
to fine-tuning. This is achieved with a re-weighing procedure. A comparative study is then provided between four state-of-the-art methods: Multisource Tradaboost\cite{tradaboost}, Task Tradaboost \cite{tradaboost}, Multi-KT \cite{svm}, transfer learning decision forests \cite{tldf}, and Weighted Multisource Tradaboost. This study is evaluted on a subset of four classes of the Caltech-256 Dataset \cite{caltech}. In turn, one of each of the four classes used is the target, while the other are used as sources. The classes chosen are dog,  horse,  leopard  and  zebra, chosen for empirically possessing a positive relationship with each other. The results show that our method can overcome the other methods in accuracy performance, but the higher asymptote assumption is still not achieved. This assumption states that a method employing transfer learning should outperform machine learning methods without transfer even when target data is abundant \cite{torrey}.

The contributions described in this paper are a comparative study exhibiting results not found in the literature, stating that the higher asymptote behavior theorized in \cite{torrey} is not achieved by several state-of-the-art methods, and a novel approach for transfer learning that addresses this limitation of the methods studied. Although this limitation is not surpassed the method is showing a way to improve transfer learning approaches towards the theoretical asymptotic performance that can be applied in several transfer learning methods.

The rest of this document is organized as follows: Section \ref{sec:transferlearning} describes necessary concepts and notation introduced by transfer Learning. Section \ref{sec:comparedmethods} describes the methods studied in this document. Section \ref{sec:design} describes the experiment ran, and the results obtained are discussed in Section \ref{sec:results}. Finally, our conclusions and possible future research directions are presented in Section \ref{sec:conclusion}.

\section{Transfer Learning}
\label{sec:transferlearning}
Transfer learning introduces several new concepts to machine learning. The definitions and notation described here will be used throughout this paper.

Standard machine learning tries to learn and then classify using one dataset for training and another one for testing. Both these datasets are assumed to come from the same distribution. In transfer learning information is leveraged from additional sources. The dataset that has the same distribution as the test data is called the target, and other(s) is(are) called source(s).

In this paper, the methods studied assume all datasets lie in the same feature space. This is called homogeneous transfer learning. If the feature space is different for at least one of the sources it is heterogeneous transfer learning (see MultiK-KT in \cite{svm}).

The success of transfer learning hinges on the inherent relationship between target and sources. In the case of a weak/non-existent connection between source and target the final classifier may actually be worse than its no-transfer counterpart. This phenomenon is known as negative transfer.

\begin{figure}[h]
	\centering
	\includegraphics{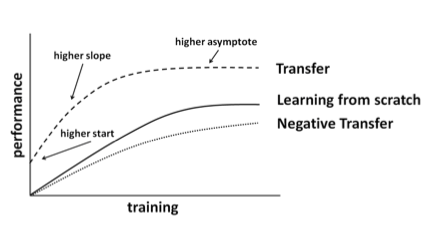}
	\caption{Three different ways in which transfer learning may improve traditional machine learning as a function of the number of target training samples: Higher Start, Higher Slope and Higher Asymptote (See Sec. \ref{sec:transferlearning}.) Use of sources with no relation with the target may lead to the behavior described by the Negative Transfer curve. Figure adapted from \cite{svm}.}
	\label{fig:transferlearning}
\end{figure}

There are three measures by which transfer may improve learning \cite{torrey} (see Fig. \ref{fig:transferlearning}): Higher Start (better performance at the beginning of learning since source information is leveraged), Higher Slope (using the transferred knowledge the new task can be learned faster), and Higher Asymptote (since more information is being leveraged, the final system should have better performance). As shall be seen by the study presented in this paper, the Higher Asymptote hypothesis doesn't always hold true, even when a positive relationship between source and target can be established (See Section \ref{sec:results}).

Finally, there is one more distinction between two types of transfer: instance transfer and task transfer. Instance transfer refers to scenarios in which some of the source data can actually be used to help train the new model. MultiSourceTradaboost is an example of this scenario. In task transfer, the source tasks are described explicitly by models trained \textit{a priori}. Multi-KT and TaskTradaboost are examples of this type of transfer.

\section{State-Of-The-Art}
\label{sec:comparedmethods}
The methods compared in this paper comprise the recent state-of-the-art approached used in low data transfer learning. The methods are now described in detail.

\subsection{Multi-KT - Support Vector Machines \cite{svm}}

In 2014, Tommasi \textit{et al.} \cite{svm} proposed a formulation for transfer learning using Support Vector Machines (SVM). Their problem setting was as follows: Assume that $j$ old (source) models (described by $\hat{w}_j$) are available \textit{a priori}, and that these models can be expressed as a weighted sum of kernel functions (\textit{e.g.} obtained \textit{a priori} from an off-the-shelf SVM package). Then, in order to leverage the information already encoded in the other models, a simple framework is presented: change the cost function of an SVM solver to include a term imposing "model fidelity" (\textit{i.e.} the cost function of the new model $w$ must be close to a weighted sum of the pre-existing j models) (see equation \ref{eq:svmhomogeneous}).

\begin{equation}
\begin{aligned}
& \underset{w,b, \xi}{\text{argmin}}
& & \frac{1}{2} \left\Vert w - \sum_{j=1}^{J} \beta_{j} \hat{w}_j\right\Vert^{2} + C \sum_{i=1}^{n} \zeta_i \xi_i^{2}\\
& \text{subject to}
& & y_i (w^T \phi (x_i) - b) \geq 1 - \xi_i,\\
& & & \xi_i \geq 0
\label{eq:svmhomogeneous}
\end{aligned}
\end{equation}
where $\zeta_i$ are used to balance the contribution of positive and negative samples, taking into account their proportion in the training set, $\beta_i$ are real numbers that control the influence that each \textit{old} model should have over the new model (estimated \textit{via} minimizing the leave-one-out error). The rest are components that make up a standard SVM formulation(\textit{i.e.} $\left\Vert w \right\Vert$ ensures margin maximization, $C\sum_{i=1}^{n} \zeta_i \xi_i^{2}$ encodes the trade-off between model fidelity and margin maximization and data fidelity; and the constraints ensure data fidelity).

\subsection{Transfer Learning Decision Forests (TLDF)\cite{tldf}}
\label{subsubsec:TLDF}

\begin{figure}[ht]
	\centering
	\includegraphics[scale = 0.5]{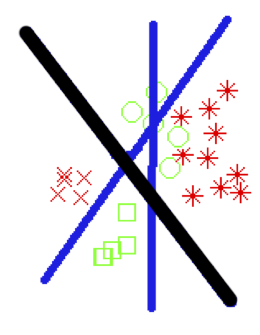}
	\caption{Consider two tasks (red and green) , each with two labels (stars and crosses for the red task, circles and squares for the green task. The red task is the target task, the green task is a source task. All three hyperplanes shown in the figure separate the target (red) dataset perfectly. The hyperplane represented in black, however, separates all the data from all the datasets simultaneously. According to the thinking presented in \cite{tldf} the black hyperplane is preferable, and should be a better minimizer of the generalisation error. Image adapted from \cite{tldf}.}
	\label{fig:splits}
\end{figure}

In 2014, Goussies \textit{et al.}\cite{tldf}, proposed a method to do transfer learning using random decision forests. This is a method that uses data from several sources to shape the decision regions.  Considering N+1 classification tasks, $T_0, \dots, T_N$, the goal is to solve the classification task $T_0$, called the target task, using the knowledge of all tasks. By leveraging information from all datasets at once, the regions generated by the decision splits of each tree in the forest will construct a classifier with a higher classification accuracy, since more information is taken into account when shaping the decision regions (see Figure \ref{fig:splits}).

Goussies \textit{et al.} go on to propose a mixed information gain formulation that formalizes the intuition described. For the k-th split:
\begin{equation}
\begin{aligned}
\theta^*_k = \underset{\theta_k} {\text{argmax}} (1-\gamma) \mathcal{I}_0(\theta_k) + \gamma \sum^{N}_{n=1} \mathcal{I}_n(\theta_k)
\label{eq:tldfsplit}
\end{aligned}
\end{equation}

\noindent where $\mathcal{I}_0$ is the information gain on the target dataset (that stems from split $\theta_k$) and $\mathcal{I}_n, n=1,\dots,N$ are the information gains on the source datasets (stemming from the same split). $\gamma$ is a trade-off parameter that regulates the importance given to the information gain on the source and target datasets.

From this formulation a new problem arises: the leaf nodes of the tree are not required to have any datapoint belonging to the target. So, after creating a tree, a label propagation procedure is applied. For a given leaf node without a single target datapoint a distance vector is constructed with the distance from that node to all other leaf nodes that have at least one target datapoint. Then, the prediction made by the closest leaf node possessing at least one target datapoint is copied to the current node without target datapoints. The distance measure used is a Mahalanobis distance using the estimated mean and estimated covariance of the leaf nodes involved.

\subsection{TrAdaboost}

In 2007, Dai \textit{et al.} proposed TrAdaboost\cite{dai}, a transfer learning variant of AdaBoost. In 2010, Yao and Doretto\cite{tradaboost}, proposed two boosting models that perform transfer learning from multiple sources: MultiSource TrAdaboost and TaskTradaboost. Adaboost works as follows: At each iteration a weak classifier is trained. Then, the samples in the training set are re-weighted, increasing the weight of misclassified samples. This forces the next weak classifier trained to focus on getting the misclassified samples right. As such, expert models are being created for all the regions of the feature space of the dataset. Then, a final classifier is constructed by weighted majority voting of all the weak classifiers. The extensions proposed by Yao and Doretto in Tradaboost included:

\subsubsection{MultiSource TrAdaboost \cite{tradaboost}}
\label{subsubsec:mstradaboost}

\begin{figure}[h]
	\centering
	\includegraphics[width = 0.45\textwidth]{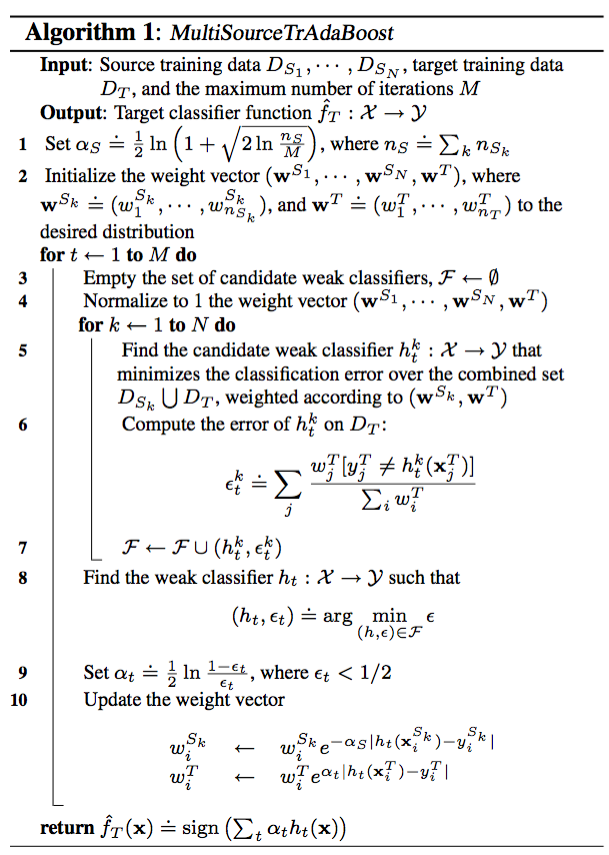}
	\caption{MultiSource Tradaboost algorithm. Taken from \cite{tradaboost}}
	\label{fig:mstradaboost}
\end{figure}

For the MultiSource TrAdaboost model, proposed by Yao and Doretto\cite{tradaboost} the availability of a very small target dataset is complemented by the availability of several larger datasets to be used as source. Information for all the datasets is leveraged by multiplexing between datasets in each iteration. When training one of the weak classifiers to boost, the target dataset is complemented by the source dataset that appears to be the most closely related to the target (\textit{i.e.} the one that leads the weak classifier to the lowest error in the target dataset in the current iteration). Then, the weights of the datapoints in all the datasets are readjusted. However, unlike in Adaboost where misclassified points have their weight increased, the re-weighting procedure differs depending on which dataset is being used. Points in the target dataset have their weight increased if they are misclassified. On the other hand, misclassified points in any source dataset have their weight reduced. This is to express the belief that if a point in a source dataset is presenting conflicting information with the target dataset,then transfer from that datapoint should be avoided. The precise algorithm used is shown in Fig.  \ref{fig:mstradaboost}, taken from \cite{tradaboost}.

\subsubsection{Task TrAdaboost \cite{tradaboost}}
The TaskTrAdaboost performs transfer from previously available models, instead of from other datasets. It is divided in two phases.

Phase I consists of training off-the-shelf Adaboost models on each of the source datasets available.

Phase II mimicks Adaboost by boosting several weak classifiers on a weighted dataset. However, in TaskTradaboost the weak models used are the Adaboost models trained on the source datasets. The weight update step in this algorithm is identical to the one in Adaboost. 

\subsubsection{Weighted Multisource Tradaboost}
\label{subsubsec:wmstradaboost}
When using transfer learning, information from both target and source datasets is leveraged. Naturally, most strategies have some way to weigh the data according to the prior belief of how similar the target data's and the source data's distribution is (\textit{i.e.:} the $\beta_j$ in Multi-KT, the $\gamma$ parameter in TLDF's, and the weight vectors $w_i^{S_k}$ and  $w_i^{T}$ in Tradaboost). However, to our knowledge, no method incorporates the proportion of target and source data available as prior knowledge in the mixing of target and source information in the learning stage.

We believe this approach to be sound because, if more target data is available, the contribution of the source data used in the model should gradually shift from model defining to fine-tuning. We postulate this is the case because as more target data becomes available, the model built using only target data becomes more and more robust. In that case, forcing the model to acommodate source data can actually be detrimental to the model's performance. We shall prove this with a comparative study in the results section.

Our approach follows the general method described by tradaboost but replaces the weight update rules defined in step 10 of the algorithm (see Fig. \ref{fig:mstradaboost}) to take into account the proportion of target and source data available.

Instead, we propose:

\begin{equation}
\begin{aligned}
w_i^{S_k} \leftarrow w_i^{S_k}e^{-\eta \alpha_S |h_t(x_i^{S_k}) - y_i^{S_k}|} \\
or \\
w_i^{S_k} \leftarrow w_i^{S_k}\eta e^{-\alpha_S |h_t(x_i^{S_k}) - y_i^{S_k}|}
\end{aligned}
\label{eq:wupdate1}
\end{equation}
\begin{equation}
\begin{aligned}
w_i^{T} \leftarrow w_i^{T}e^{\eta \alpha_T |h_t(x_i^{T}) - y_i^{T}|}\\
or\\
w_i^{T} \leftarrow w_i^{T}\eta e^{\alpha_T |h_t(x_i^{T}) - y_i^{T}|}
\end{aligned}
\label{eq:wupdate2}
\end{equation}

where $\eta$ is a term that depends on the amount of target and source data available for training. The same term can be used for both target and source datapoints because the weight update step (shown in equations \ref{eq:wupdate1} and \ref{eq:wupdate2})inverts the signal of the exponent when switching dataset. Strategies for how to define this quantity are discussed in Sec. \ref{sec:design}.
\section{Experimental Design}
\label{sec:design}
We compare all the methods described in Section \ref{sec:comparedmethods} with a subset of the Caltech-256 Dataset \cite{caltech}. This is a dataset composed of 256 classes, with images as datapoints. The images range from high-quality pictures to poor drawings of the subject of the class. A subset of 4 classes was chosen from those available: dog, horse, leopard and zebra as well as the background class. These classes were shown to test positive transfer from empirically related classes: 4-legged animals. For these classes we downloaded the Scale Invariant Feature Transform (SIFT) features from \cite{features} (See \cite{features} for details). These features have a dimension of 300.

The results presented are averaged over 5 tests done with random permutations of the data. For each test, the results are averaged over 4 runs, each with a different 4-legged animal as target. As such, all experiments are averaged over twenty runs. Finally, the tests are run with the number of target points available ranging from 1 to 10.

For comparison with the no-transfer scenario, an off-the-shelf SVM classifier is trained exclusively on the target data.

\subsection{Method Hyperparameters}
\label{subsec:hyperparameters}
For Multi-KT the $C$ parameter (see Equation \ref{eq:svmhomogeneous}) is chosen via cross-validation on the source data. The $\beta_j$ parameters are chosen by minimizing the leave-one-out error. In \cite{svm} feature fusion is used. For fairness of comparison with the other methods only SIFT features were used.

For the Random Forests methods the parameters were decided according to the values found in the literature instead of chosen by testing different values for the parameters. This was due to the long time needed for each run of this method. The parameters used were: $\gamma$ = 0.8 (Controls the influence of sources and target when calculating splits), Maximum tree depth = 10,  number of trees in a forest = 3.

For MultiSource Tradaboost and Task Tradaboost the only hyperparameter is the number of iterations to run. This value was set at 50 due to computational limitations.

For Weighted Multisource Tradaboost 2 different values were empirically chosen for testing: 
\begin{itemize}
\item $\eta = \frac{N_T * 100}{N_S}$
\item $\eta = \frac{N_T^2 * 100}{N_S}$
\end{itemize}
where $N_T$ is the number of target datapoints and $N_S$ is the number of source datapoints. The factor of 100 inserted in the numerator describes the belief that in most transfer learning settings target data will be scarce while source data will be abundant. So both these terms enforce that the influence of source samples will be greatly diminished as more target samples become available. Each of these values was tested on both variants shown in Equations \ref{eq:wupdate1} and \ref{eq:wupdate2}, resulting in four different tests. %Although this approach managed to have a better classification accuracy that its non-weighted counterpart, the problem of not being able to surpass no-transfer approaches when target data becomes abundant remains. More research is needed in this area.
 
For comparison with the no-transfer scenario, an off-the-shelf SVM classifier is trained exclusively on the target data. The hyperparameters for this model are the same as those used for Multi-KT but setting all the $\beta_j$ to 0.

\section{Results}
\label{sec:results}
\begin{figure}[ht]
	\centering
	\includegraphics[width = 0.5\textwidth]{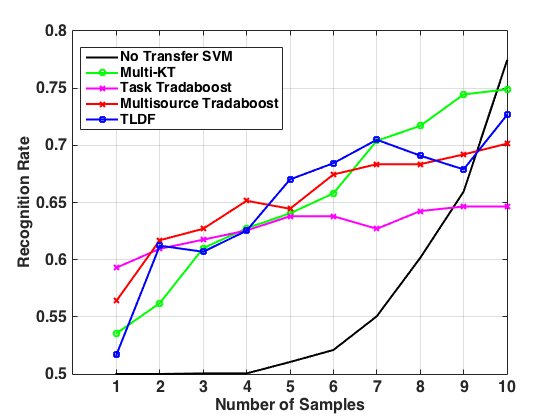}
	\caption{Results obtained running the experiment described in Section \ref{sec:design}. Each point represents the average over 20 runs with random sample and target selection.}
	\label{fig:results1}
\end{figure}

Running the comparative study of the methods described in Section \ref{sec:comparedmethods} on the Caltech-256 Dataset, the graph on Figure \ref{fig:results1} was drawn. 

As can be observed in Figure \ref{fig:results1}, the no transfer scenario outperforms all other approaches when 10 target samples are available. None of the methods studied are able to achieve the higher asymptote behaviour\cite{torrey} (see Section \ref{sec:transferlearning}).

All methods studied outperform the no-transfer approach in scenarios where the number of target samples available is very limited, up to 7 target samples.

The fact that all methods get overtaken when more target data is available suggests that once "high-quality" (target) data is available in sufficient quantity the methods are unable to extract information from the sources in a way that is not conflicting with the targets. This implies that further protection from negative transfer is required.

The TLDF method shows very unstable performance. Also, results have been found in the literature stating that this method outperforms no-transfer in cases where more target data is available. Only 1-10 target datapoints are available in our experiment, and this amount of data is not enough to populate a feature space with a dimension of 300. Since the feature space is sparsely populated, during the label propagation step the distances between leaf nodes with no target datapoints and the closest leaf node with a target can be immense, which could justify the instability found.

\begin{figure}[ht]
	\centering
	\includegraphics[width = 0.5\textwidth]{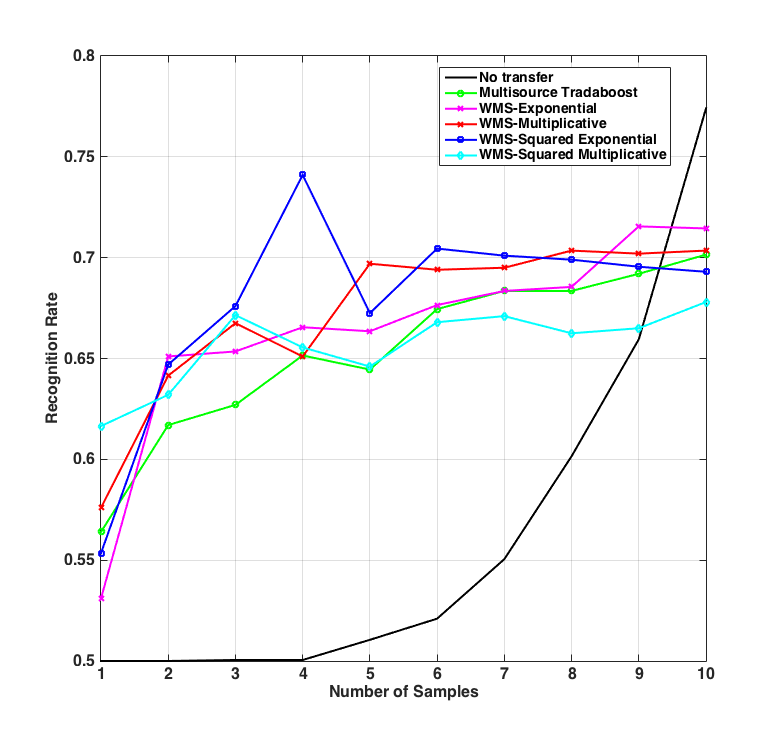}
	\caption{Results obtained running the experiment described in Section \ref{sec:design} for Weighted Multisource Tradaboost. Each point represents the average over 20 runs with random sample and target selection.}
	\label{fig:results2}
\end{figure}

To address this limitation of failing to achieve the higher asymptote behaviour, the Weighted Multisource Tradaboost method was applied to the same dataset in the same conditions. The results obtained are shown in Figure \ref{fig:results2}.

In this figure, WMS-Exponential and WMS - Multiplicative refer to using the linear $\eta$ weight shown in Section \ref{subsec:hyperparameters} witht the weight update rules defined in Equations \ref{eq:wupdate1} and \ref{eq:wupdate2} respectively. WMS-Squared exponential and WMS-Squared Multiplicative correspond to the same weight update rules using the squared $\eta$ weight shown in Section \ref{subsec:hyperparameters}.
As can be seen in Figure \ref{fig:results2} all attempts outperform Multisource Tradaboost except for the Squared Outside attempt. However, the failure to achieve the higher asymptote behavior still eludes us. More research is needed on this topic.

\section{Conclusion}
\label{sec:conclusion}

All methods studied outperform the no-transfer approach when very little target data is available only to get outperformed by it when more target data is accessible. This failure to achieve the higher asymptote behaviour theorized in \cite{torrey} is an unpublished result, and is one of the contributions of this work. Our proposed attempt to improve Multisource Tradaboost to achieve the higher asymptote behaviour did not come to fruition, albeit managing to improve the classification performance. Further research is required to determine when to expect the no-transfer approach to be preferable, and how to achieve the higher asymptote. The strategies employed in this paper can be ported to other transfer learning methods, namely the Multi-KT method. Our strategy to improve transfer learning approaches towards the theoretical asymptotic performance predicted in \cite{torrey} is our other contribution.

\subsection{Future Work}
According to the results obtained the following future research is suggested:
\begin{itemize}
    \item Add a regularization term in the $\beta_j$ calculation steps for Multi-KT that takes into account how much target data is available. This would hopefully lead the method to not be outperformed by no-transfer approaches, fusing the best of both worlds
    \item Further testing with the Random Forests approaches is needed in order to evaluate the performance of these methods in situations where more target data is available. Results found in the literature indicate that this research is promising.
    \item Test all the methods in a scenario where the classes are unrelated and test specifically for resilience against negative transfer
\end{itemize}
{%\small
%\bibliographystyle{ieee}
%\bibliography{bibliography}

}

\end{document}